\documentclass[sigconf]{acmart}
\usepackage{algorithm}
\usepackage{algorithmic}
\usepackage{multirow}
\usepackage{bm}
\usepackage{subfigure}

\AtBeginDocument{%
  \providecommand\BibTeX{{%
    \normalfont B\kern-0.5em{\scshape i\kern-0.25em b}\kern-0.8em\TeX}}}

\copyrightyear{2021}
\acmYear{2021}
\setcopyright{acmcopyright}
\acmConference[MM '21]{Proceedings of the 29th ACM
	International Conference on Multimedia}{October 20--24, 2021}{Virtual Event, China}
\acmBooktitle{Proceedings of the 29th ACM International Conference on Multimedia
	(MM '21), October 20--24, 2021, Virtual Event, China}
\acmPrice{15.00}
\acmDOI{10.1145/3474085.3475376}
\acmISBN{978-1-4503-8651-7/21/10}

\begin{document}
\fancyhead{}

\title{Weight Evolution: Improving Deep Neural Networks Training \\
	through Evolving Inferior Weight Values}

\author{Zhenquan Lin}
\affiliation{%
	\institution{South China University of Technology}
	\streetaddress{}
	\city{}
	\country{}}
\email{zhenquan_lin@163.com}

\author{Kailing Guo}
\authornote{Corresponding author.}
\affiliation{%
  \institution{South China University of Technology}
	\streetaddress{}
  \city{}
  \country{}}
\email{guokl@scut.edu.cn}

\author{Xiaofen Xing}
\affiliation{%
	\institution{South China University of Technology}
	\institution{UBTECH-SCUT Union Laboratory}
	\streetaddress{}
	\city{}
	\country{}}
\email{xfxing@scut.edu.cn}

\author{Xiangmin Xu}
\affiliation{%
	\institution{South China University of Technology}
	\institution{Institute of Modern Industrial Technology \\of SCUT in Zhongshan}
	\streetaddress{}
	\city{}
	\country{}}
\email{xmxu@scut.edu.cn}


\begin{abstract}
To obtain good performance, convolutional neural networks are usually over-parameterized. This phenomenon has stimulated two interesting topics: pruning the unimportant weights for compression and reactivating the unimportant weights to make full use of network capability. However, current weight reactivation methods usually reactivate the entire filters, which may not be precise enough. Looking back in history,  the prosperity of filter pruning is mainly due to its friendliness to hardware implementation, but pruning at a finer structure level, i.e., weight elements, usually leads to better network performance. We study the problem of weight element reactivation in this paper. Motivated by evolution, we select the unimportant filters and update their unimportant elements by combining them with the important elements of important filters, just like gene crossover to produce better offspring, and the proposed method is called weight evolution (WE). WE is mainly composed of four strategies. We propose a global selection strategy and a local selection strategy and combine them to locate the unimportant filters. A forward matching strategy is proposed to find the matched important filters and a crossover strategy is proposed to utilize the important elements of the important filters for updating unimportant filters. WE is plug-in to existing network architectures. Comprehensive experiments show that WE outperforms the other reactivation methods and plug-in training methods with typical convolutional neural networks, especially lightweight networks. Our code is available at https://github.com/BZQLin/Weight-evolution.
\end{abstract}




\maketitle

\section{Introduction}
Deep neural networks have achieved state-of-the-art results in various machine learning tasks \cite{DBLP:journals/pami/ChenPKMY18, DBLP:conf/nips/KrizhevskySH12, DBLP:conf/cvpr/HeZRS16,DBLP:conf/cvpr/XieLHL20,DBLP:conf/iccv/DingGDH19}. One dominant reason for this success is the innovations in network architecture, e.g., shortcut \cite{DBLP:conf/cvpr/HeZRS16} and batch normalization \cite{DBLP:conf/icml/IoffeS15}. Although the novel network architectures achieve excellent results, over-parameterization \cite{DBLP:conf/iclr/0022KDSG17, DBLP:conf/nips/DenilSDRF13, DBLP:journals/corr/HanMD15}is still the inherent problem in deep neural networks.

Network pruning \cite{DBLP:conf/cvpr/GuoWLY20,DBLP:conf/iclr/LiuSZHD19,DBLP:conf/nips/Dong019,DBLP:conf/icml/TanM20} is one kind of the typical methods to solve the problem of over-parameterization. It tends to compress a network effectively by cutting out the unimportant or redundant parts of the network without significant loss of performance. Network pruning can be roughly divided into weight pruning and filter pruning. Weight pruning \cite{DBLP:conf/iclr/FrankleC19,DBLP:conf/cvpr/KwonLKKPW20} focuses on pruning individual weights and achieves a very high theoretical compression rate, but it is unfriendly to hardware implementation to achieve real compression. Filter pruning takes a filter as an integrated part for removing. Since it does not need specialized implementation for real speed-up, most recent network pruning methods only focus on filter pruning \cite{DBLP:conf/iclr/0022KDSG17,DBLP:conf/ijcai/LinJZZW020,DBLP:conf/cvpr/HeDLZZ020,DBLP:conf/cvpr/LinJWZZ0020,DBLP:conf/accv/LiCLSZ20}. From another viewpoint, some recent studies \cite{DBLP:conf/cvpr/MengCLXJSL20,DBLP:conf/cvpr/PrakashSFZ19,DBLP:conf/icml/JeongS19} try to reactivate the pruned components of the network during training to improve performance. Possibly because of the prosperity of filter pruning, the recent reactivation methods are operated on entire filers. However, when looking back to the history of network pruning, weight pruning usually achieves higher classification accuracy with the same level of compression rate, which indicates that finer structure leads to better performance. Since the aim of reactivation methods is to improve network performance other than compression, they can drop the burden of implementation for real acceleration. However, weight element-wise reactivation is overlooked in previous works. In this paper, we conjecture that a finer pruning structure serves as a better starting point for final network performance in network reactivation.

To reactivate weight elements for performance improvement, two problems need to be addressed: how to select the weight elements to be reactivated and how to set new values to them. Evolution is a process of improving the environment adaptiveness of a specie by making good genes become dominant and eliminating bad genes. Motivated by this, we select the unimportant weights (good genes) and update them with important weights (bad genes). We name our method weight evolution (WE). By analogy, we treat a filter as an individual and the weight elements as genes. Borrowing the terminologies from evolution, we name unimportant filters/weights as inferior filters/weights and important filters/weights as dominant filters/weights. In the remaining, we abuse these items. 

\begin{figure*}[t]
	\includegraphics[width=1\textwidth,height=0.24\textheight]{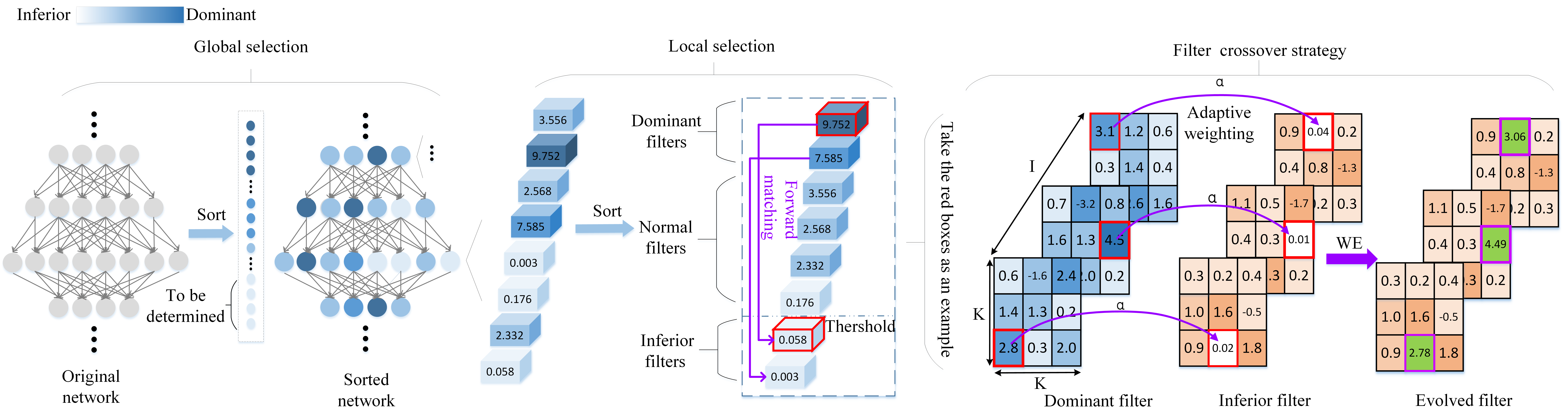}
	\caption{An illustration of the procedure of weight evolution}
	\label{fig:evolution}
\end{figure*}

A filter's importance can be roughly determined by its $\ell_{1}$ norm, but there are some layers in which most filters have large or small $\ell_{1}$ norms. Thus, we also need the relative importance of a filter in the specific layer for the final judgment. In this paper, we successively use global and local criteria for double-checking to select the inferior filters. We forwardly match the inferior filters with dominant filters according to their importance order. Finally, the crossover strategy updates the values of inferior elements in the inferior filters by summing with dominant weight values in dominant filters with adaptive coefficient. Different from the method 
re-initializing and pruning (RePr) \cite{DBLP:conf/cvpr/PrakashSFZ19}
that reinitialize the pruned filters to be orthogonal to the un-pruned filters, dominant values in dominant filters are inherited to update the values of inferior elements. WE is different from filter grafting (FG) \cite{DBLP:conf/cvpr/MengCLXJSL20} that combines the whole dominant filter and inferior filter. A sudden change of the whole filter may make the training unstable. Instead, our evolution method only updates a few elements of inferior filters and thus is more stable for training. Figure \ref{fig:evolution} illustrates the procedure of weight evolution.

WE only adds little extra computation and can be easily added to each iteration or epoch of training. This underlines the universality and practicality of WE. We conduct comprehensive experiments on the image classification task. The results show that the weight evolution method consistently improves the performance of existing neural networks, especially on lightweight network architectures such as MobileNetV2. We also provide ablation experiments to study the effectiveness of the four strategies in WE and the superiority of element level reactivation over filter level reactivation, and show that WE is compatible to both batch normalization (BN) and convolutional layers.

\section{Related Work}
{\bf Network pruning}:
The purpose of network pruning is to cut off the redundant or unimportant parts in neural networks to reduce parameters and accelerates reasoning speed. 
Weight pruning \cite{DBLP:journals/corr/HanMD15, DBLP:conf/eccv/ZhangYZTWFW18, DBLP:conf/cvpr/TungM18, DBLP:conf/nips/DingDZGHL19} and filter pruning \cite{DBLP:conf/iclr/0022KDSG17, DBLP:journals/tcyb/HeDKFYY20,DBLP:conf/iccv/HeZS17, DBLP:journals/corr/HuPTT16,  DBLP:conf/iccv/LiuLSHYZ17, DBLP:conf/iccv/LuoWL17, DBLP:conf/iclr/YeL0W18, DBLP:conf/nips/ZhuangTZLGWHZ18}are two common network pruning categories. 

Weight pruning is to prune specific weight elements in the network. Han et al. \cite{DBLP:journals/corr/HanMD15} prunes weight elements with small magnitudes and retrain the pruned network to retain the performance. Zhang et al. \cite{DBLP:conf/eccv/ZhangYZTWFW18} takes weight pruning as an optimization problem by constraining the number of weight elements less than a threshold, and solves the problem with alternating direction method of multipliers, but retraining is still needed. To avoid retraining, Ding et al. \cite{DBLP:conf/nips/DingDZGHL19} modifies momentum stochastic gradient descent (SGD) by sparsifying the gradients when updating. Although these methods achieve good theoretical compression rate and speed-up, they result in unstructured network architectures and need sophisticated hardware or software to achieve real acceleration.

In contrast, filter pruning prunes the whole filters in neural networks. The key problem filter pruning is how to define pruning criterion. ``Smaller-norm-less-important'' is an intuitive principle for pruning. The $\ell_{1}$ norm \cite{DBLP:conf/iclr/0022KDSG17} and the $\ell_{2}$ norm \cite{DBLP:journals/tcyb/HeDKFYY20} have been used as pruning criterion. Channel pruning \cite{DBLP:conf/iccv/HeZS17} and ThiNet \cite{DBLP:conf/iccv/LuoWL17} determine the channels to be pruned by minimizing reconstruction error. Considering the scaling factors in BN layer indicating the importance of the corresponding convolutional filters, the works \cite{DBLP:conf/iccv/LiuLSHYZ17} and \cite{DBLP:conf/iclr/YeL0W18} prune the unimportant filters by sparsifying the scaling factors of BN layers. Zhuang et al \cite{DBLP:conf/nips/ZhuangTZLGWHZ18} utilize extra discriminative information to guide the pruning process. Network trimming \cite{DBLP:journals/corr/HuPTT16} proposes to prune zero activation neurons since the activation layer's output influences the following convolutional layer. Filter pruning is easy to implement with existing hardware and software, and it becomes the most popular network pruning method in recent years.

Different from network pruning that compresses and accelerates the model, this paper studies the problem of reactivating filters to improve the network performance.

{\bf Filter restoration}: Restoring/reactivating unimportant or redundant filters in the neural network can effectively improve the performance of the neural network without changing the structure of the network. RePr \cite{DBLP:conf/cvpr/PrakashSFZ19} trains the subnetwork after pruning the redundant filters. The pruned filters are then restored to be orthogonal to the remaining ones and the entire network is retrained. Filter grafting \cite{DBLP:conf/cvpr/MengCLXJSL20} reactivate filters with small $\ell_{1}$ norm by adaptively balancing the grafted information among internal filters and networks. Qiao et al. \cite{DBLP:conf/cvpr/QiaoLZY19} rejuvenates neurons by improving the utilization of computational resources, which detects dead filters and calculates the utilization of resources on-the-fly. Due to the popularity of filter pruning, element level reactivation is lacked of consideration. In this paper, we improve the network performance from the perspective of reactivating individual elements.

{\bf Other utilization strategies}:
To improve the utilization of computational resources, another think way is to train a network that is hard to be compressed. Some recent methods impose orthogonality constraints to the weights of a network to reduce redundancy during training.
Singular value bounding \cite{DBLP:conf/cvpr/JiaTGX17,DBLP:journals/pami/LiJWLT21} proposes a training scheme that forces the singular values of weight matrices of each layer to be near one after every epoch.
Orthogonalization method using Newton’s iteration (ONI) \cite{DBLP:conf/cvpr/Huang00WYL020} utilizes proxy matrices that can be transformed into orthogonal matrices as the parameters for training and the transformation is conducted through Newton's iteration. Wang et al. \cite{wang2020orthogonal} impose filter orthogonality on a convolutional layer based on the doubly block-Toeplitz matrix representation of the convolutional
kernel and derive a new regularization term in the loss function for pursuing orthogonality.
These methods can work with existing architectures to improve performance. From another insight, retrospective loss (RL) \cite{DBLP:conf/kdd/JandialCSGKB20} proposes to utilize the past model state during the training process to guide the current training. RL serves as an extra item for the loss function for performance improvement.

\section{Weight evolution}
Inspired by the evolution, we try to reactivate unimportant weights with important weight values during network training, just like good genes replacing bad genes, to improve network performance in this paper.

In this section, global and local selection criteria are first proposed to identify inferior filters. Then, we introduce the forward matching and crossover strategies to find  and inherit information from the corresponding dominant filters. Detailed training scheme and computational complexity analysis are also given.

\subsection{Global and local selection criteria}
\begin{figure}[t]
	\centering
	\includegraphics[width=0.46\textwidth,height=0.33\textheight]{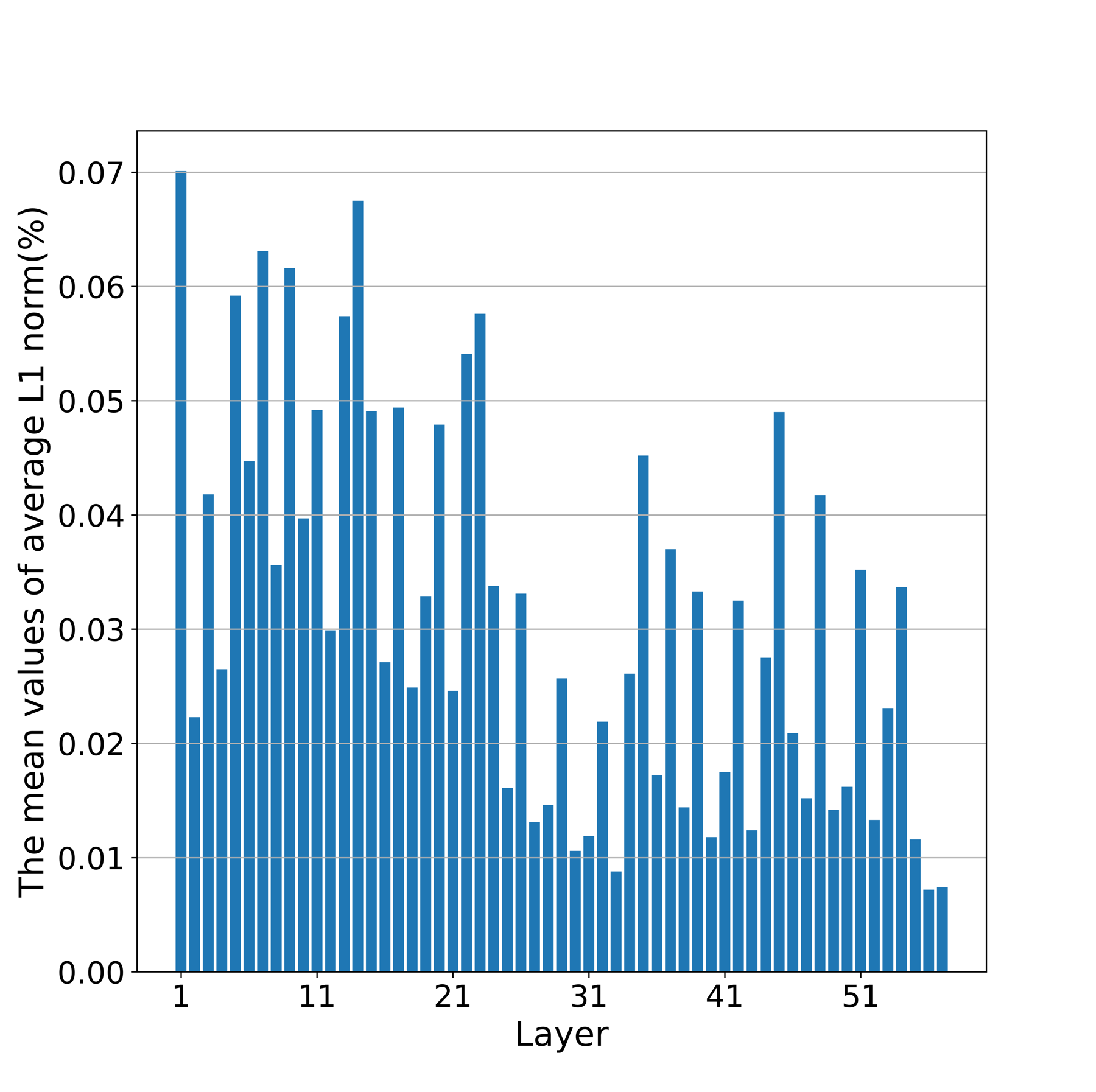}\\
	\caption{The histogram of the mean values of average $\ell_{1}$ norm}
	\label{fig:hist}
\end{figure}

We need to identify the inferior filters before reactivation. The $\ell_{1}$ and $\ell_{2}$ norms are simple pruning criteria in previous filter pruning methods \cite{DBLP:conf/iclr/YeL0W18, DBLP:journals/tcyb/HeDKFYY20}. Here we choose the $\ell_{1}$ norm due to its computational simplicity. Since the number of filter channels varies in different layers, it is necessary to reduce the influence of channel numbers. We compute the average $\ell_{1}$ norm to measure the importance of a filter. However, the distributions of average $\ell_{1}$ norms of filters in different layers are very different. Taking MobileNetV2 \cite{DBLP:conf/cvpr/SandlerHZZC18} on CIFAR-10 \cite{krizhevsky2009learning} as an example, we plot the mean values of average $\ell_{1}$ norms in each convolutional layer in Figure \ref{fig:hist}. We can see that the mean values are varies significantly. Thus, we need to consider the importance of a filter both globally and locally.

\subsubsection{Global selection}
We use $\mathcal{W}_{i,j}\in \mathbb{R}^{I\times K\times K}$ to denote the $j$-$th$ filter of the $i$-$th$ layer of the neural network. Here, $I$ denotes the input channels and $K$ denotes the kernel size. The average $\ell_{1}$ norm of a filter is computed as follows:
\begin{equation}\label{avg_l1_norm}
	||\mathcal{W}_{i,j}||_{1}^{avg} = \frac{\sum\limits_{c=1}^I \sum\limits_{k_1=1}^K\sum\limits_{k_2=1}^K |\ \mathcal{W}_{i,j}(c,k_1,k_2) |\ }{I}
\end{equation}
We sort the filters according to their average $\ell_{1}$ norms and select the smallest ones temporarily for further determination with a selection rate $r$. We refer to these temporary filters as filters to be determined, denoted with a set $\mathbb{W}^{tbd}$.

In this paper, we use the popular step learning rate decay method for training. It is found that the classification accuracy of neural networks fluctuates greatly at the beginning of training or when the learning rate suddenly decreases. This is because the network needs to be trained for several iterations to become stable when it comes to a new area of the solution space. In addition, we empirically find that the average $\ell_{1}$ norms are closer to each other at the beginning of sudden learning rate decrease. Thus, the selection rate should be small at the beginning of a training stage. We enumerate two conditions which the selection rate $r$ needs to satisfy:
\begin{itemize}
	\item The selection rate $r$ should be small at the beginning of a training stage, gradually increase, and become stable as the network converges.
	\item When the learning rate suddenly decreases, $r$ also needs to decrease.
\end{itemize}

\begin{figure}[t]
	\centering
	\includegraphics[width=0.5\textwidth,height=0.27\textheight]{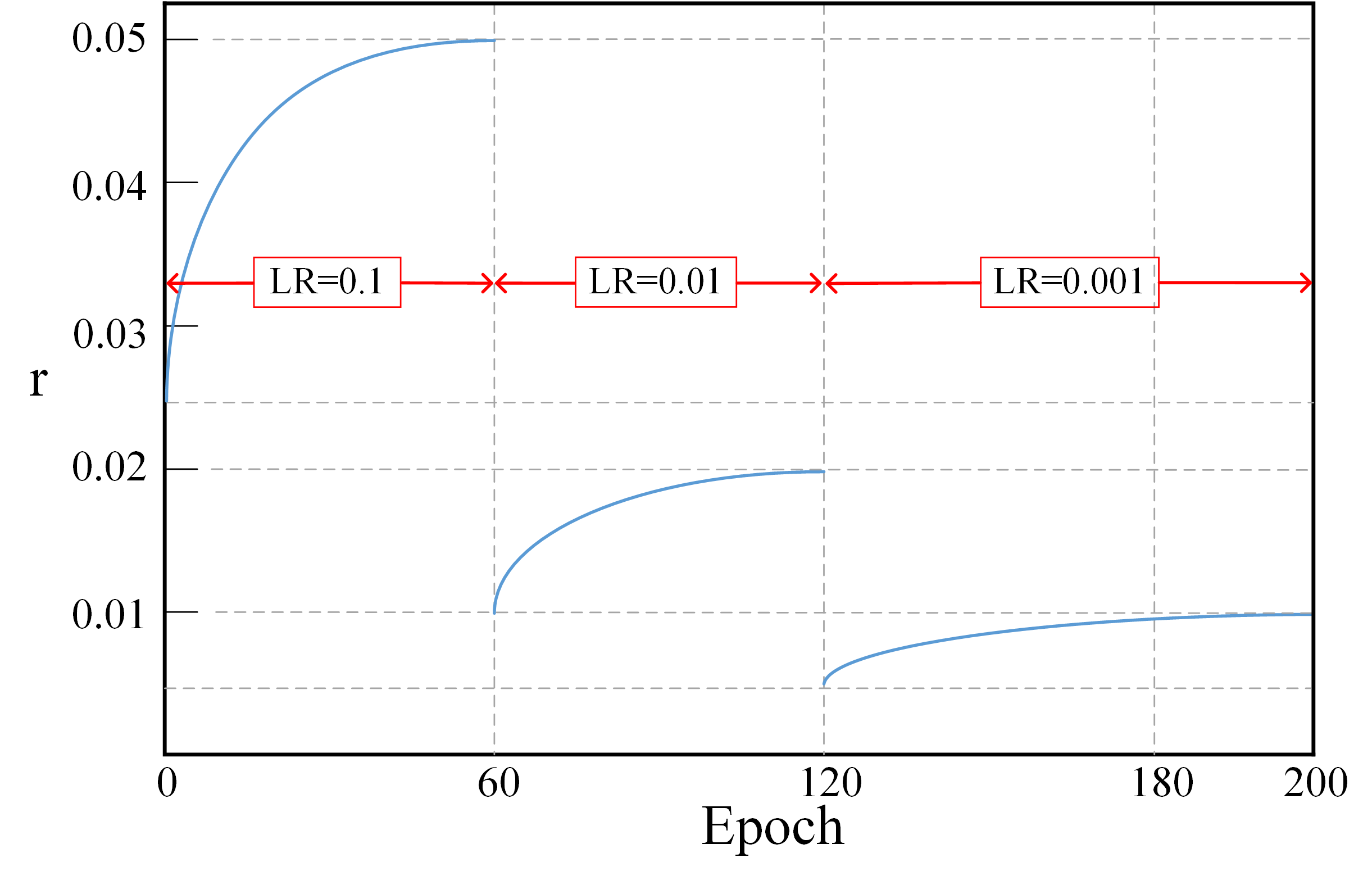}\\
	\caption{The change of global selection rate $r$}
	\label{fig:env}
\end{figure}

The setting function of $r$ for the $e$-$th$ epoch at the $t$-th decay stage is given by
\begin{equation}\label{global_selecton_rate}
	r(e)=(\hat{r}/\beta^{t-1})\text{sigmoid}((e-e_{0})/\eta),
\end{equation}
where $e_0$ denotes the starting epoch of the current stage, and $\hat{r}$, $\beta$, and $\eta$ are parameters needed to be set. The sigmoid function is defined as
\begin{equation}\label{sigmoid_function}
	\text{sigmoid}(x)=1/(1+e^{-x}).
\end{equation}
Here we choose the sigmoid function because it gradually increases from 0.5 to 1 when $x\geq 0$, which is in line with the requirements of the changing trend of the selection rate. The parameter $\beta$ is larger than one to make sure that $r$ is smaller in the later stages. This is because when the model is well trained, all filters are more reliable. The parameter $\eta$ is set to make sure that the sigmoid function at the end of the current decay stage is near one.
Taking 200-epoch training with step-decay learning rate, the curve of $r$ is illustrated in \ref{fig:env}. Details of the setting are given in Section \ref{experiment}.

\subsubsection{Local selection}
After global selection, we can know the layers where the filters to be determined reside. We then locally select the final inferior filters in these layers according to relative importance in the specific layer. The relative importance of a filter $\mathcal{W}$ is defined as
\begin{equation}\label{RI}
	RI(\mathcal{W})=||\mathcal{W}||_1/||\mathcal{W}^{dom}||_1,
\end{equation}
where $\mathcal{W}^{dom}$ is the most dominant filter (i.e., the one with largest $\ell_{1}$ norm) located in the same layer as $\mathcal{W}$. If the relative importance of a filter to be determined is less than a predefined threshold $\gamma$, the corresponding filter is chosen as the final inferior filter. By combining global and local selection, we can identify inferior filters without mistaking some relative dominant filters but with small average norms.

\subsection{Filter matching and crossover}
After local selection, we first match the inferior filters with dominant filters and then upgrade the inferior filters with the good genes from the corresponding dominant filters.

\subsubsection{Forward matching strategy}
Suppose the number of inferior filters is $c$. We use
$\mathbb{W}^{inf,i}=\{\mathcal{W}^{inf,i}_{1}, \cdots, \mathcal{W}^{inf,i}_{c}\}$ to denote the set of inferior filters in the $i$-$th$ layer, where filters in the set are sorted in ascending order according to their $\ell_{1}$ norms. Without misunderstanding, we drop the superscript $i$ for simplification in the following. The inferior filter set  $\mathbb{W}^{inf,i}$ is simplified into $\mathbb{W}^{inf}=\{\mathcal{W}^{inf}_{1}, \cdots, \mathcal{W}^{inf}_{c}\}$. Correspondingly, we select the $c$ largest $\ell_{1}$ norm filters in the $i$-$th$ layer and sort them in ascending order to construct the set $\mathbb{W}^{dom}=\{\mathcal{W}^{dom}_{1}, \cdots, \mathcal{W}^{dom}_{c}\}$ of dominant filters.

\begin{figure}[t]
	\centering
	\includegraphics[width=0.44\textwidth,height=0.31\textheight]{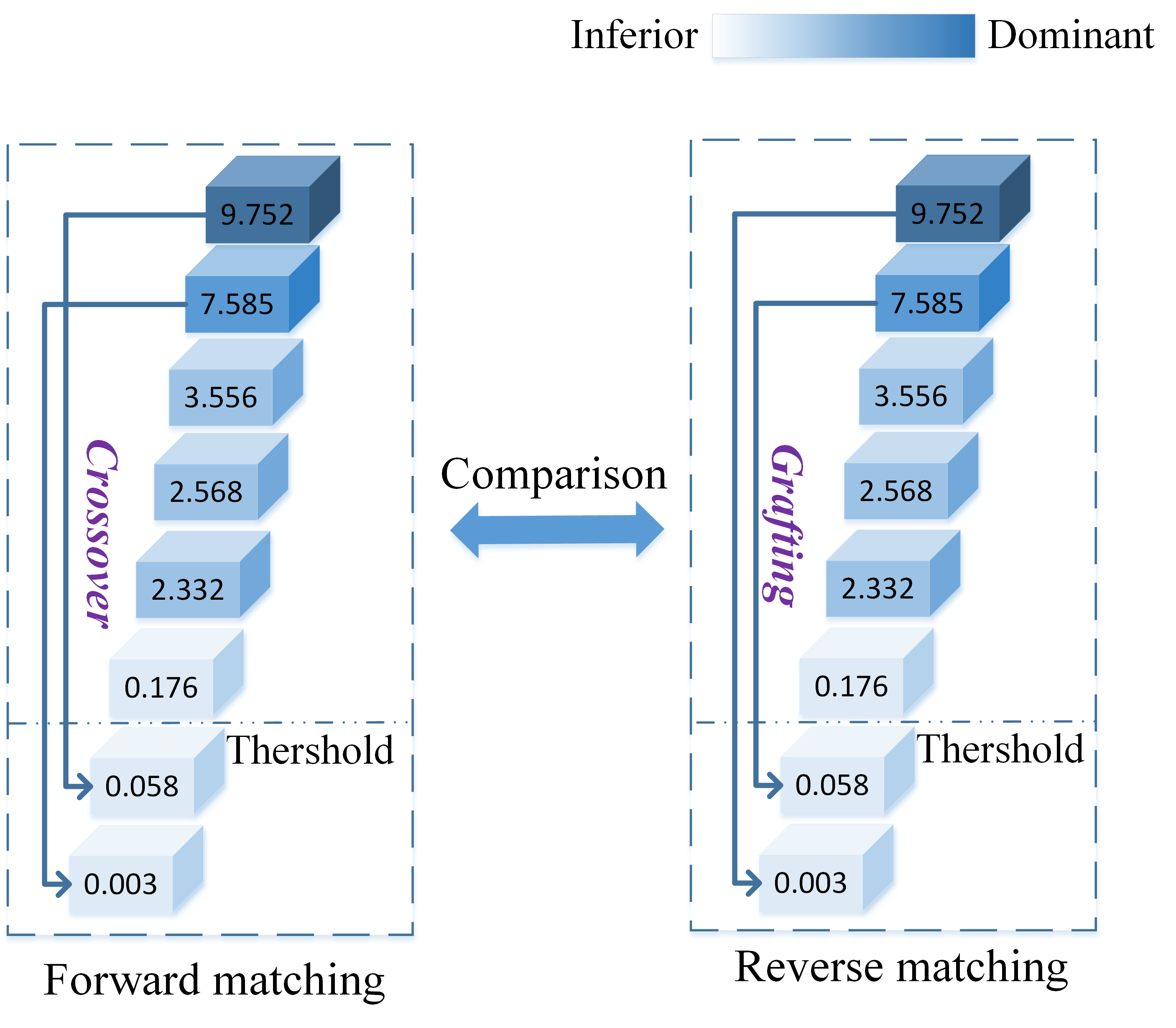}\\
	\caption{The comparison of the reverse and forward matching strategies}
	\label{forward_matching}
\end{figure}

We need to decide the matching order of these two sets. When replacing inferior filters with dominant filters, FG \cite{DBLP:conf/cvpr/MengCLXJSL20} uses a reverse matching strategy:
matching $\mathcal{W}^{inf}_{1}$ with $\mathcal{W}^{dom}_{c}$, $\mathcal{W}^{inf}_{2}$ with $\mathcal{W}^{dom}_{c-1}$, and so on.
However, by using such strategy, the most inferior filter in the dominant filters is matched with the most dominant filter in the inferior filters, and their gap maybe not distinguishable enough. In this paper, we propose a forward matching strategy that keeps the order of filter importance, i.e.,
we match $\mathcal{W}^{inf}_{1}$ with $\mathcal{W}^{dom}_{1}$, $\mathcal{W}^{inf}_{2}$ with $\mathcal{W}^{dom}_{2}$,
and so on. Figure \ref{forward_matching} illustrates the comparison of the reverse and forward matching strategies. We will show that forward matching works better than reverse matching in the experiment section.
\subsubsection{Crossover strategy}
In evolution, crossover means mixing the genes of two individuals. In our method, we use crossover to denote mixing the weight elements of two filters. We use $W^{dom}_{d,s}$ and $W^{inf}_{d,s}\in \mathbb{R}^{K \times K}$ to denote the $s$-$th$ slice of the $d$-$th$ filter in the sets $\mathbb{W}^{dom}$ and $\mathbb{W}^{inf}$, respectively. As shown in Figure \ref{fig:evolution}, we update the inferior filters slice-to-slice. This is because each slice of a filter works as an individual feature extractor of the corresponding input channel. For simplification, we reshape slices into vectors and drop the subscipts $d$ and $s$, i.e., $W^{dom}_{d,s}$ and $W^{inf}_{d,s}$ change into $\mathbf{w}^{dom}$ and $\mathbf{w}^{inf}\in \mathbb{R}^{K^{2}}$. We use $w_{k}$ to denote the $k$-$th$ value of a vector $\mathbf{w}$. The index of dominant gene of  $\mathbf{w}^{dom}$ is given by $p=\mathop{\arg\max}_{k}\,|w_{k}^{dom}|$, and the index of the inferior gene of $\mathbf{w}^{inf}$ is given by $q=\mathop{\arg\min}_{k}\,|w_{k}^{inf}|$. The crossover operation is to replace $w_{q}^{inf}$ with
\begin{equation}\label{element_update}
	\hat{w}_{q}^{inf}=\alpha w_{q}^{inf}+ (1-\alpha) w_{p}^{dom},
\end{equation}
where $\alpha$ is a coefficient to weight the two genes. We adaptively set the value of $\alpha$ according to the gene relative importance as follows:
\begin{equation}\label{alpha}
	\alpha=\frac{|w_{q}^{inf}|}{|w_{q}^{inf}|+|w_{p}^{dom}|}
\end{equation}
Since $|w_{p}^{dom}|$ is always larger than $|w_{q}^{inf}|$, we can transfer more information from the dominant gene to the offspring.

\subsection{Training Scheme}
Here we show how to combine weight evolution with existing SGD training method.
After every training epoch, we first determine the inferior filters through global and local selection. Then, we update the inferior elements in the inferior filters with forwarding matching and crossover strategies and go through to the next training epoch.

We derive the weight evolution procedure from ordinary convolution, but it is also suitable for depth-wise convolution and $1\times 1$ convolution. This is because the weight of an ordinary convolution filter is a third-order tensor, and the weights of a depth-wise convolution and a $1\times1$ convolution filter are respectively matrix and vector, i.e., special case of third-order tensor. WE is also applicable to BN layer because it can be considered as applying an extreme case of depth-wise convolution (kernel size of 1) after normalization. In this case, each filter is a scalar. The bias of BN or convolution is also a scalar and can be applied with WE. When applying WE to group convolution, the relative importance is computed in the specific group because the information in a different group is relatively independent. In the following, we abuse the term filter for all kinds of convolution and BN.

The training scheme is summarized in Algorithm \ref{alg}. Note that we set the update interval to be one epoch in this paper, but it can be changed to iteration numbers and tuned to pursue further improvement.

\subsection{Computational Complexity}
Our method adds some extra operations at the end of each epoch.
The training cost increases but in a limited amount. We analyze the extra computational complexity of our method and other related training methods.
Suppose the output and input channel numbers and kernel size of a layer are $N$, $I$, and $K$, respectively.
The computation amounts of all the compared methods mainly come from multiplication. We give their computational complexities of multiplication as follows: WE: $O(NIK^{2})$, FG: $O(NIK^{2})$, RePr: $O(N^{2}IK^{2})$, SVB: $O(N^{2}IK^{2})$. 
WE is of the same order as FG and is less than RePr and SVB. Note that the complexities of updating operations in WE and FG are low-order items and dropped. Suppose the proposition of inferior filters in WE and FG are both $r$. The dropped complexity items are given by $O(rNI)$ and $O(rNIK^{2})$ for WE and FG, respectively. We can see that the actual amount of multiplication operations of WE is smaller than that of FG. Here we just analyze FG's complexity for single model. When dealing with multiple model, its complexity will scale up with the number of models.

\begin{algorithm}[t]
	\caption{Weight Evolution Training Scheme}\label{alg}
	\begin{algorithmic}[1]
		\REQUIRE ~~\\ The weights $\mathcal{W}$ of all $L$ layers of a network, the training epochs $T$, the filter numbers of the $i$-$th$ layer $C_{i}$, the parameters for global selection $\hat{r}$, $\beta$, and $\eta$, and the relative importance threshold for local selection $\gamma$.
		\FOR{$e=1,\cdots, T$}
		\STATE Update the parameters using SGD based methods for one epoch.
		\STATE Sort the filters across the network by average $\ell_{1}$ norm given in Eq. (\ref{avg_l1_norm}).
		\STATE Construct the set $\mathbb{W}^{tbd}$ with filters with smallest average $\ell_{1}$ norms by the global selection rate $r$ (computed by Eq. (\ref{global_selecton_rate})).
		\FOR{$i=1, \cdots, L$}
		\STATE $\mathbb{W}^{inf}=\{\}$.
		\FOR{$j=1, \cdots, C_{i}$}
		\IF {$\mathcal{W}_{i,j}\in \mathbb{W}^{tbd}$ and $RI(\mathcal{W}_{i,j})<\gamma$ (computed by Eq. (\ref{RI}))}
		\STATE put it into the set $\mathbb{W}^{inf}$.
		\ENDIF
		\ENDFOR
		\STATE Sort the filters in $\mathbb{W}^{inf}$ in ascending order according their $\ell_{1}$ norms.
		\STATE Count the filter numbers of $\mathbb{W}^{inf}$, denoted by $c$.
		\STATE From the set $\mathbb{W}^{dom}$ by selecting the $c$ largest $\ell_{1}$ norm filters in the $i$-$th$ layer and sorting them in ascending order.
		\FOR{$j=1, \cdots, c$}
		\STATE Update $\mathcal{W}_{j}^{inf}$ with $\mathcal{W}_{j}^{dom}$ by Eqs. (\ref{element_update}) and (\ref{alpha}).
		\ENDFOR
		\ENDFOR
		\ENDFOR
	\end{algorithmic}
\end{algorithm}
\section{Experiment}\label{experiment}

In this section, we evaluate WE and other comparison methods on benchmark image classification datasets: CIFAR-10 \cite{krizhevsky2009learning}, CIFAR-100 \cite{krizhevsky2009learning}, and  ImageNet \cite{DBLP:journals/ijcv/RussakovskyDSKS15}. Several representative popular network architectures are taken into account, including ResNet \cite{DBLP:conf/cvpr/HeZRS16}, DenseNet \cite{DBLP:conf/cvpr/HuangLMW17}, MobileNetV1 \cite{DBLP:journals/corr/HowardZCKWWAA17}, MobileNetV2 \cite{DBLP:conf/cvpr/SandlerHZZC18}, and ShuffleNet \cite{DBLP:conf/cvpr/ZhangZLS18}. Comprehensive experiments are conducted with Pytorch \cite{ketkar2017introduction} to show the effectiveness of WE.

\subsection{Datasets and Experiment settings}
\subsubsection{Datasets}
Both CIFAR datasets contain 60,000 color images with the size of $32\times 32$.
The images in the CIFAR-10 dataset are divided into 10 classes and each class has 5,000 training images and 1,000 testing images. CIFAR-100 has 100 classes of images and there are 500 training images and 100 testing images per class. Following standard data augmentation \cite{DBLP:conf/eccv/ZeilerF14, DBLP:conf/nips/KrizhevskySH12} for training, each image side is zero-padded with 4 pixels, and a $32\times 32$ sample is randomly cropped from the padded image or its horizontal flip with a probability of 0.5. For testing, we directly use the original $32\times 32$ images. Both training and testing images are
normalized with channel means and standard deviations.

ImageNet \cite{DBLP:journals/ijcv/RussakovskyDSKS15} is a large-scale image classification dataset containing approximately 1.28 million color images for training, 50,000 color images for validation. For ResNet-34 \cite{DBLP:conf/cvpr/HeZRS16} , we adopt the default ImageNet preprocessing of PyTorch for training and validation. For MobileNetV2 \cite{DBLP:conf/cvpr/SandlerHZZC18} , we use the same preprocessing, except that the network input images are resized to $128 \times 128$.

\subsubsection{Experiment settings}
For training on the CIFAR datasets, the detailed setups of the experiment are as follows: Training is based on SGD with a batch size of 128, momentum of 0.9, weight decay of 0.0005, and total training epochs of 200. Starting from 0.1, the learning rate is decreased by one-tenth at the $60$-$th$ and $120$-$th$ epochs. The experiments are run for 5 times and the mean results are reported.

The training hyper-parameters for ImageNet are consistent with the official PyTorch settings: the batch size, momentum, weight decay, and initial learning rate are set to 256, 0.9, 0.0001, and 0.1, respectively. We train the networks with SGD for 90 epochs and the learning rate is decreased by a factor of 0.1 per 30 epochs. For ImageNet, ResNet-34 \cite{DBLP:conf/cvpr/HeZRS16} is trained with two Nvidia 2080Ti GPUs, and MobileNetV2 \cite{DBLP:conf/cvpr/SandlerHZZC18} is trained with one Nvidia 2080Ti GPU.

For global selection, we set $\hat{r}$ to 0.05 for all the experiments. Specifically, we set $\beta=2.5$ and $\eta=15$ for the CIFAR datasets, and $\beta=2$ and $\eta=8$ for ImageNet. The relative importance threshold $\gamma$ for local selection is set to 0.05. Unless otherwise stated, we reactivate both convolutional layers and BN layers in the following.

\subsection{Image Classification Results}
\subsubsection{Results on CIFAR}
We conduct experiments on the CIFAR datasets with ResNet-56, ResNet-110, and DenseNet-40. We use FG-$n$ to denote the filter grafting of $n$ models. We report the classification accuracy in Table \ref{CIFAR_ordinary}. WE works better than the other methods except that it is a little worse than RePr when applied with ResNet-110. However, RePr needs QR decomposition to achieve orthogonalization for reactivation, which is also much more complex than our method. In contrast, WE only needs to reactivate the filters with a few scalar multiplication 
and adding after sorting. Besides, RePr needs to store the previous model parameters, which is also time-consuming when the model is large.

\begin{table}[h]
	\renewcommand{\arraystretch}{1.2}
	\caption{Classification accuracy (\%) comparison on CIFAR with ordinary networks
	}\label{CIFAR_ordinary}
	\scalebox{0.95}{
		\centering
		\begin{tabular}{llccc}
			\toprule
			Dataset                    & Method                & ResNet-56 & ResNet-110 & DenseNet-40 \\
			\midrule
			\multirow{8}{*}{CIFAR-10}  & Baseline  & 93.52 &93.72 &94.15  \\
										& FG-1  & 93.14         & 93.48          & 94.29             \\
										& ONI                & 93.67         &  94.30        & 94.09          \\
										& SVB        &93.16 & 93.17 &93.35           \\
										& RePr        &-- & \textbf{94.60} & 93.80           \\
										& RL   &93.48 &93.71 &--\\
										& \textbf{WE}  &\textbf{93.85} & 94.31 & \textbf{94.61}\\
			\midrule
			\multirow{7}{*}{CIFAR-100} & Baseline   & 71.78  & 71.84   & 74.54  \\
										& FG-1    & 71.00         & 72.75          & 74.56            \\
										& ONI    &71.72          &72.63          &74.24     \\
										& SVB        &71.77 & 73.36 & 73.72    \\
										& RePr                            & --   & 73.60   &74.80 \\
										& \textbf{WE}   & \textbf{72.46}  & \textbf{73.78}  &\textbf{74.99} \\
			\bottomrule
	\end{tabular}}
\end{table}

We also compare the performance of the training methods when applied with lightweight networks MobileNetV1, MobileNetV2, and ShuffleNetV1. The results are given in Table \ref{CIFAR_lightweight}. All the training methods improve the performance compared to the baseline and WE is the best. It is worth noting that the performance gains of WE for lightweight networks are more significant than ordinary networks. Specifically, after applying WE, some lightweight networks can outperform the best result of some heavyweight models even though the performance of baselines are poorer, e.g., MobileNetV2 and ResNet-110 on CIFAR-10, and ShuffleNetV1 and ResNet-56 on CIFAR-100. For fairness, WE only needs to compare filter grafting with one network. The results in Table \ref{CIFAR_lightweight} show that WE even outperforms filter grafting with 2 and 4 models (FG-2 and FG-4). This is very encouraging since FG-2 and FG-4 utilize much more information than WE.

\begin{figure*}[!htb]
	\centering
	\subfigure[ResNet-56  CIFAR-10]{\includegraphics[width=0.24\textwidth]{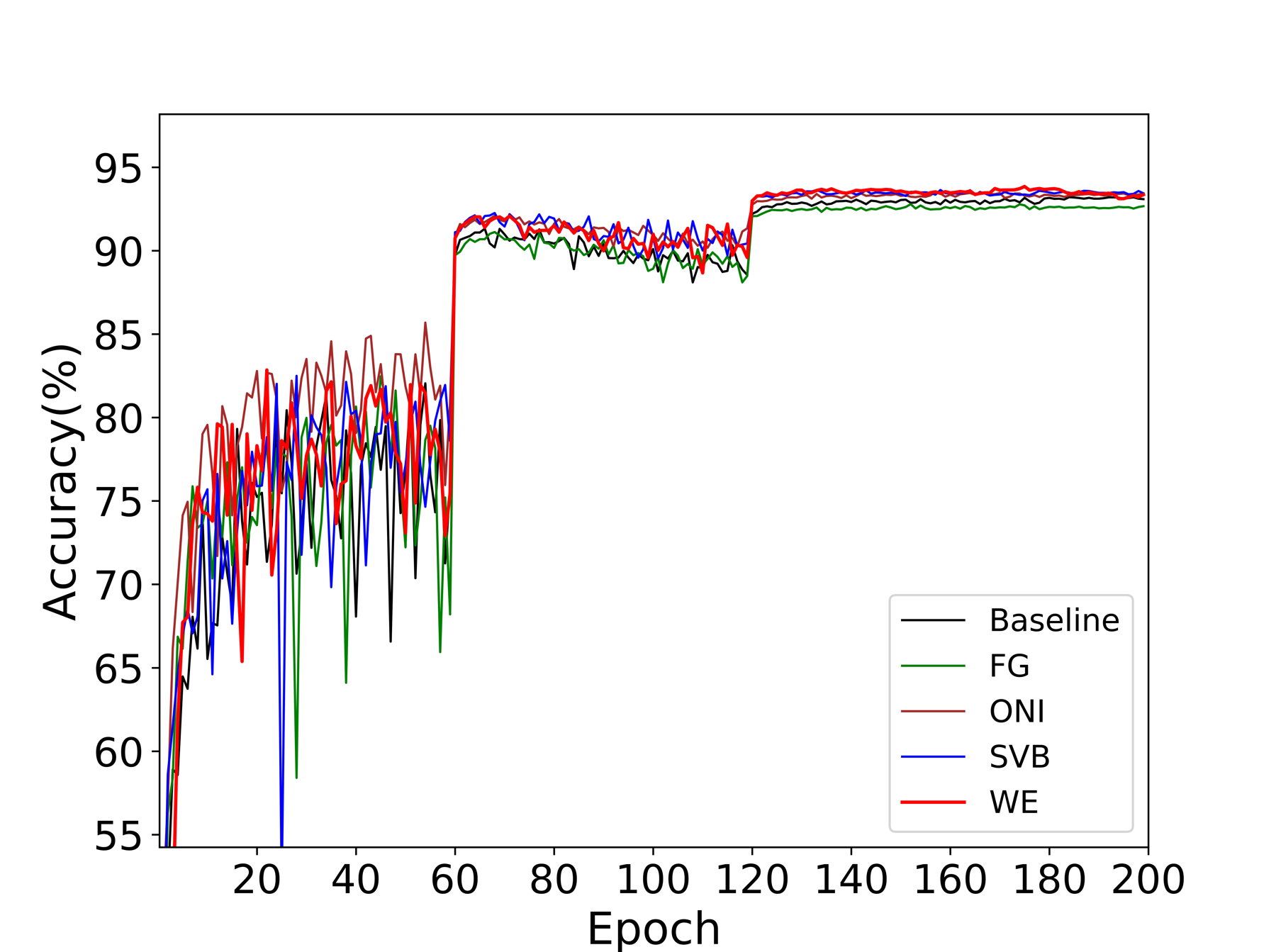}}
	\subfigure[ResNet-56  CIFAR-100]{\includegraphics[width=0.24\textwidth]{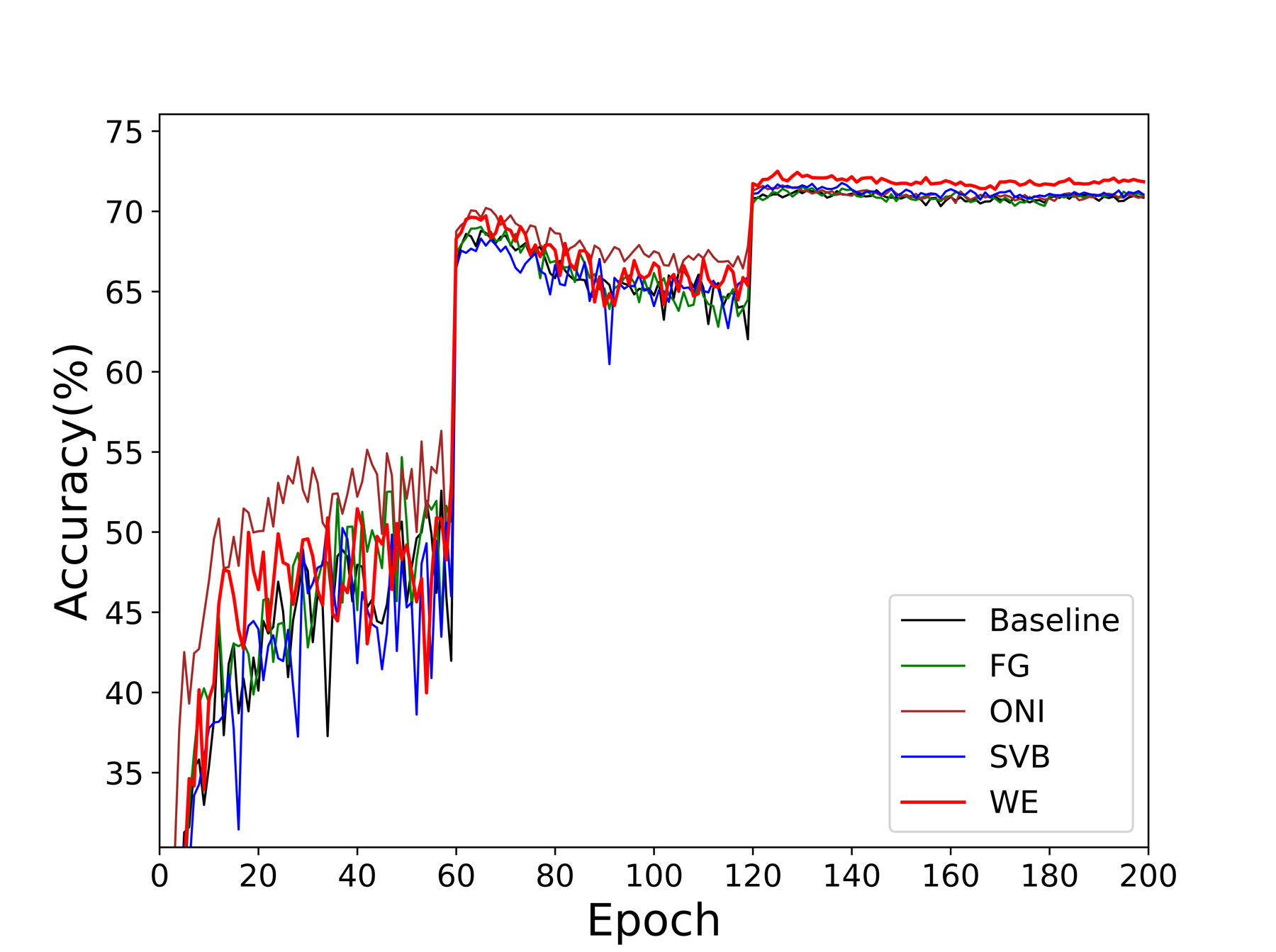}}
	\subfigure[ReNet-110  CIFAR-10]{\includegraphics[width=0.24\textwidth]{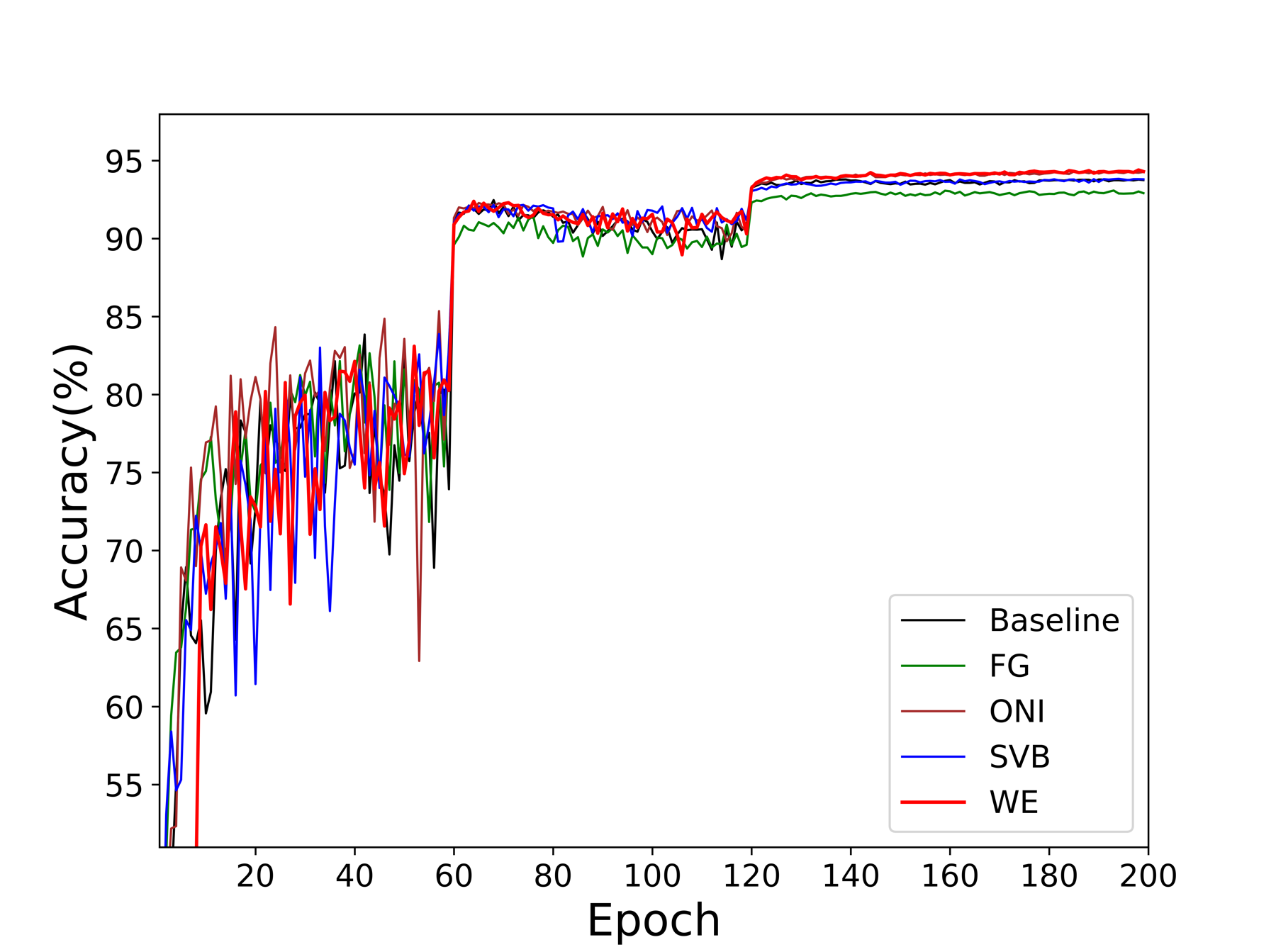}}
	\subfigure[ResNet-110  CIFAR-100]{\includegraphics[width=0.24\textwidth]{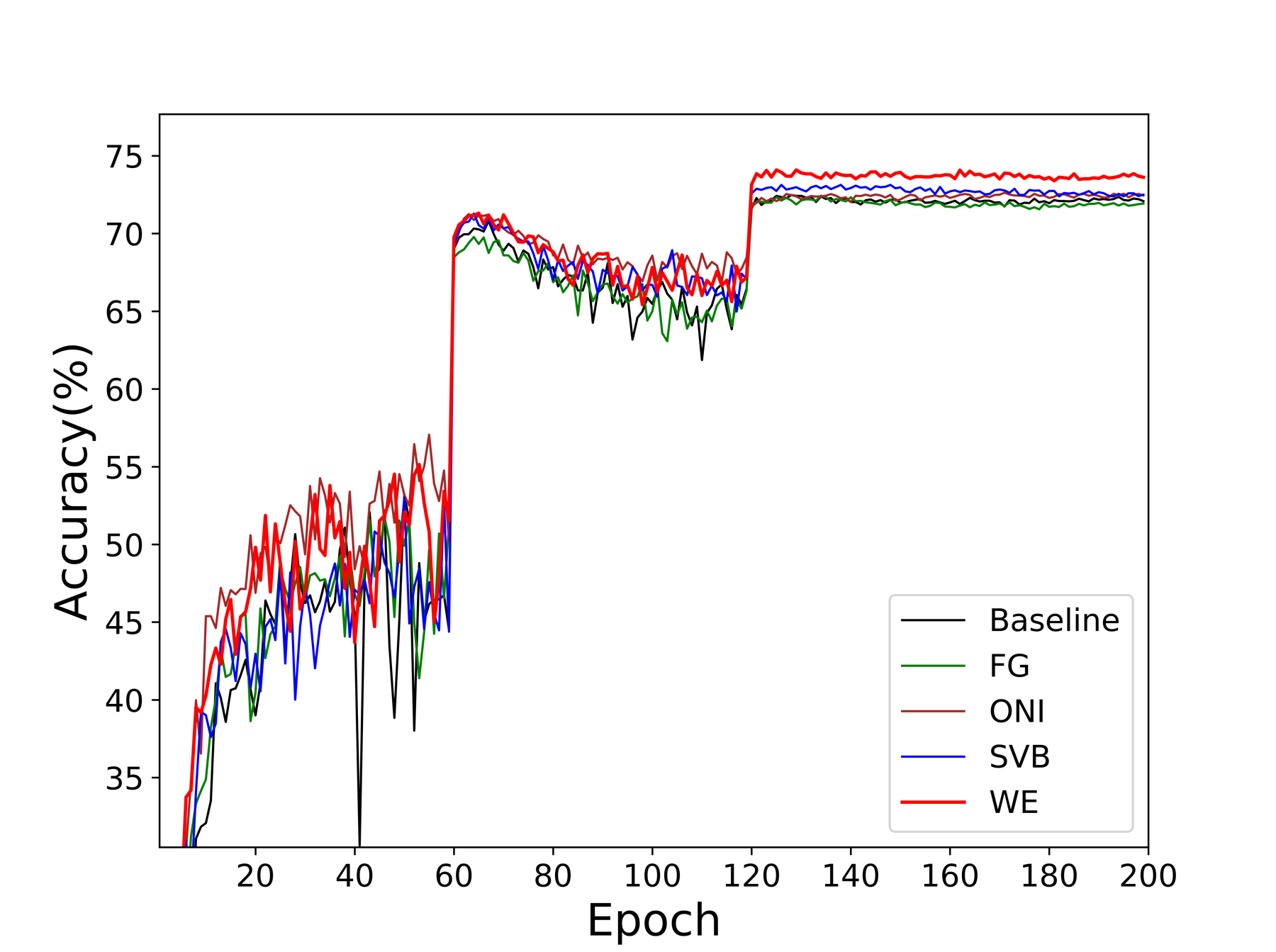}}
	\vspace{0.2in}
    \hfil
	\centering
	\subfigure[MobileNetV2  CIFAR-10]{\includegraphics[width=0.24\textwidth]{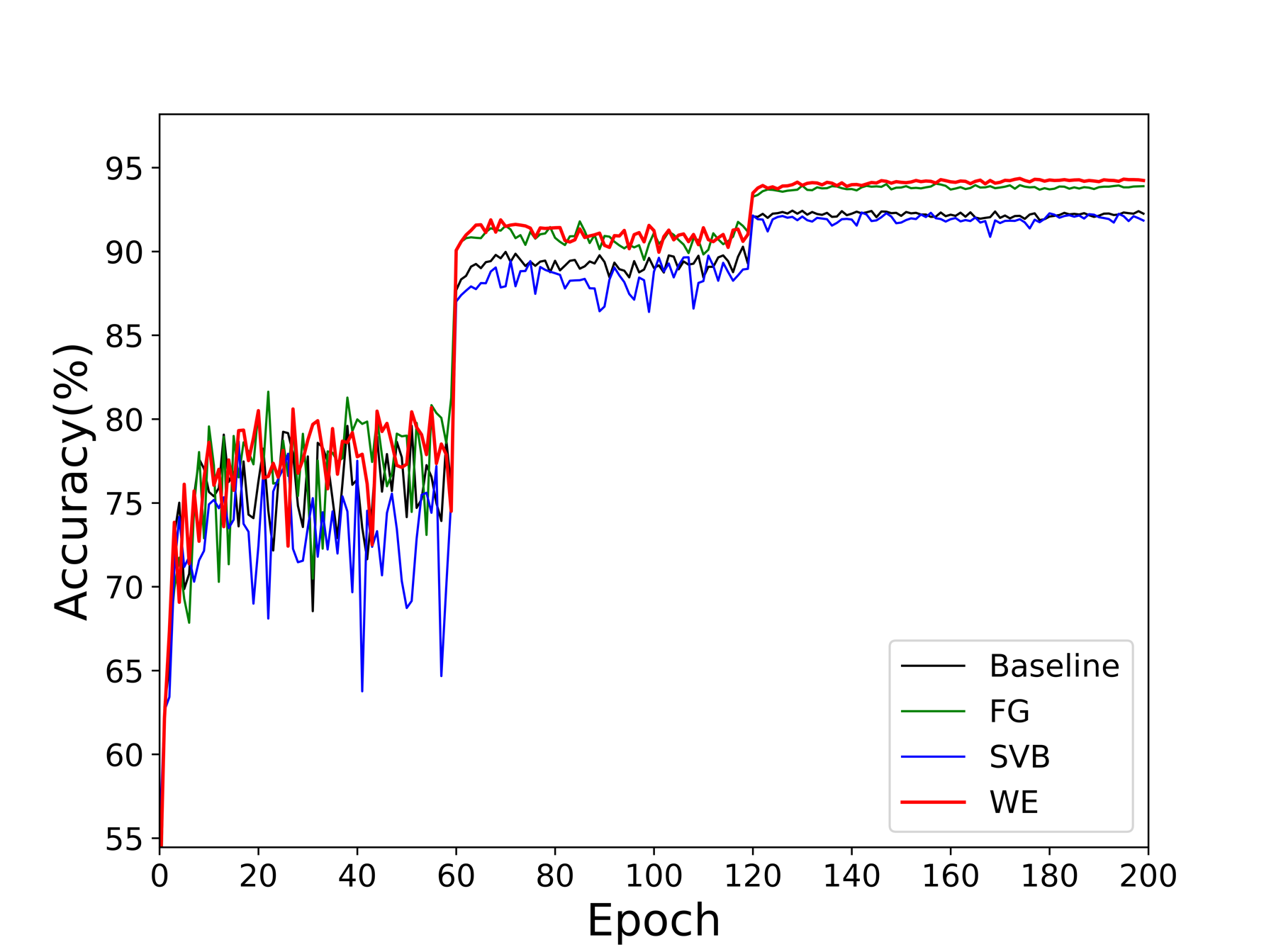}}
	\subfigure[MobileNetV2  CIFAR-100]{\includegraphics[width=0.24\textwidth]{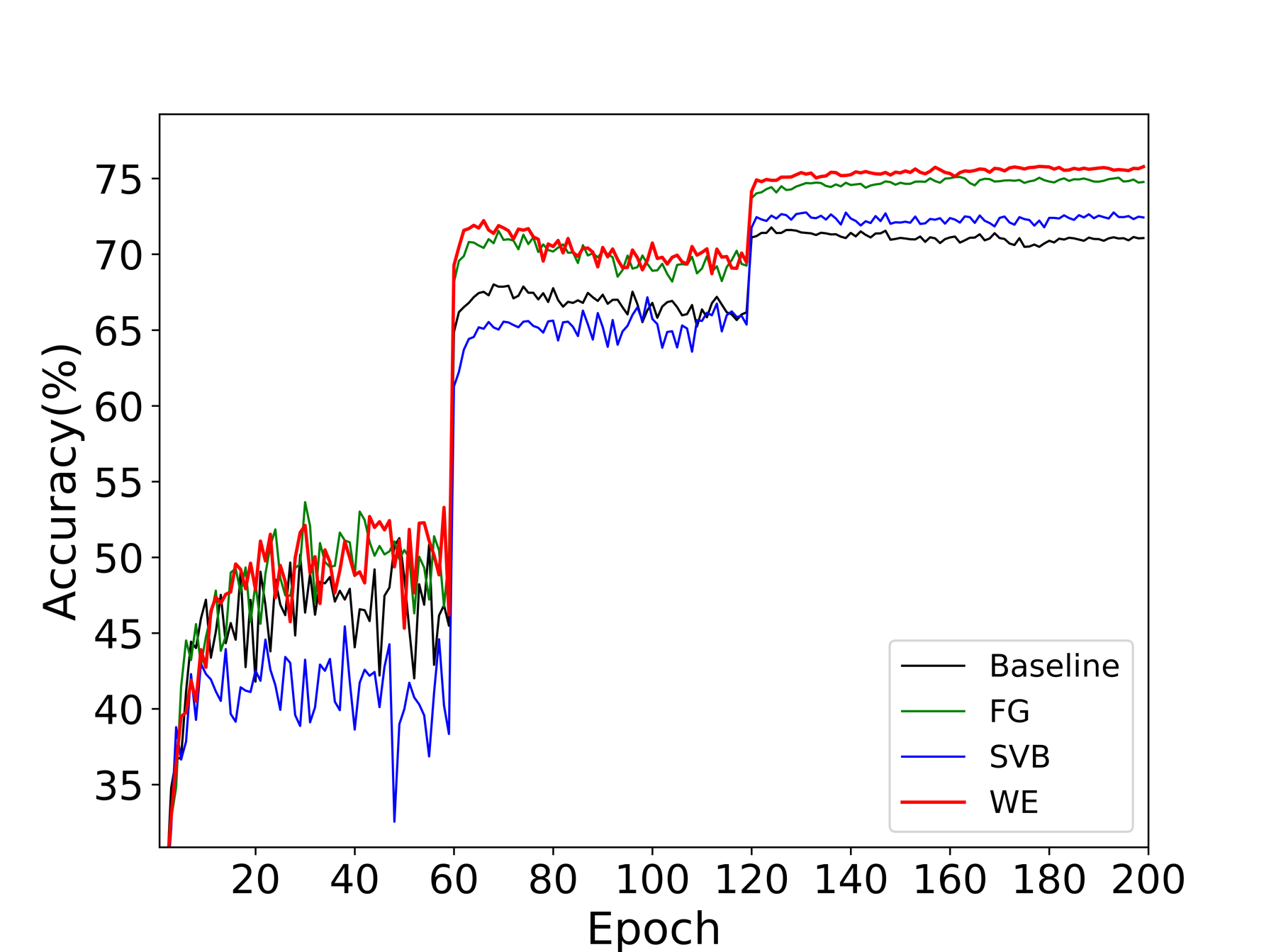}}
	\subfigure[ShuffleNetV1  CIFAR-10]{\includegraphics[width=0.24\textwidth]{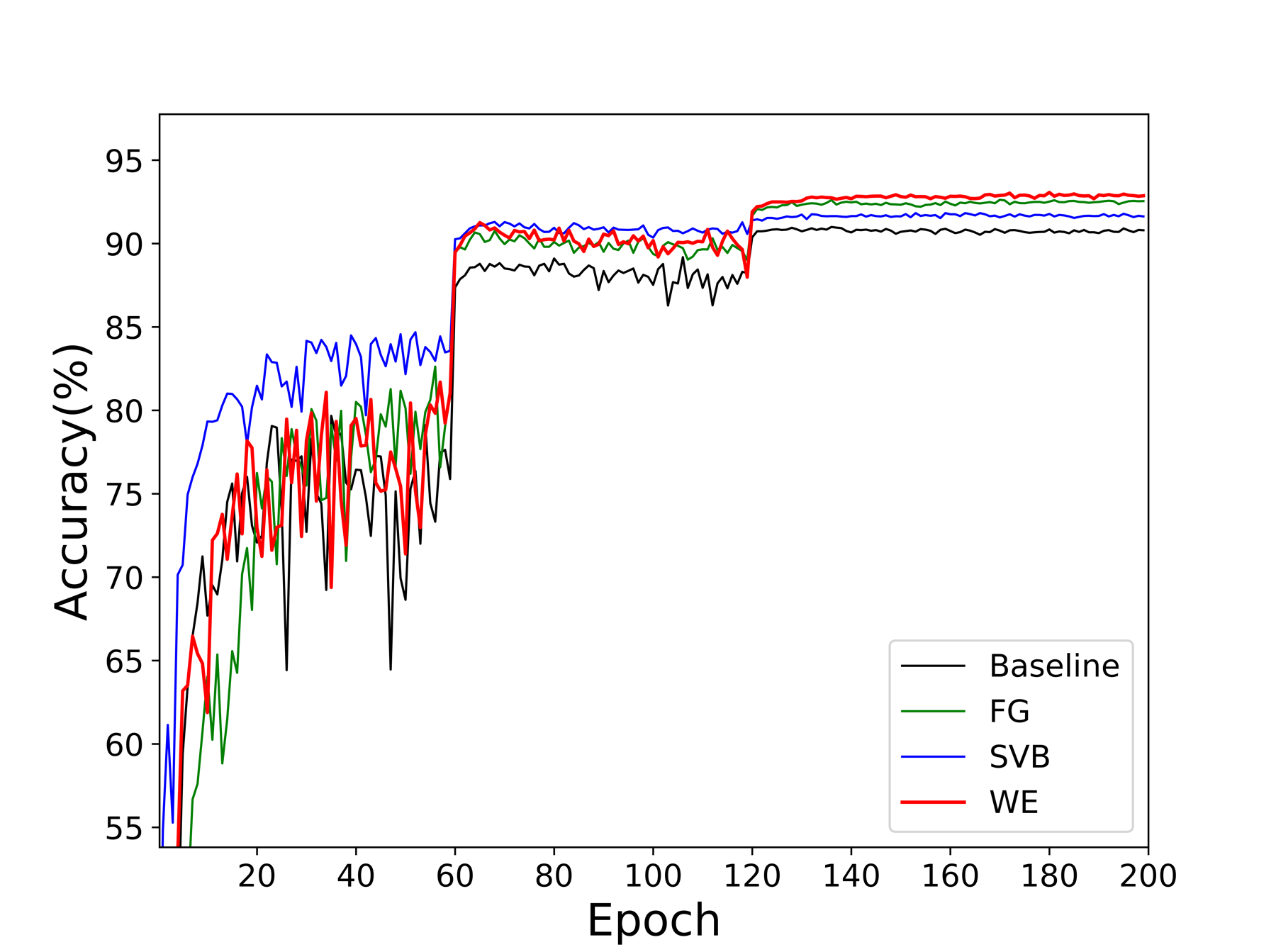}}
	\subfigure[ShuffleNetV1  CIFAR-100]{\includegraphics[width=0.24\textwidth]{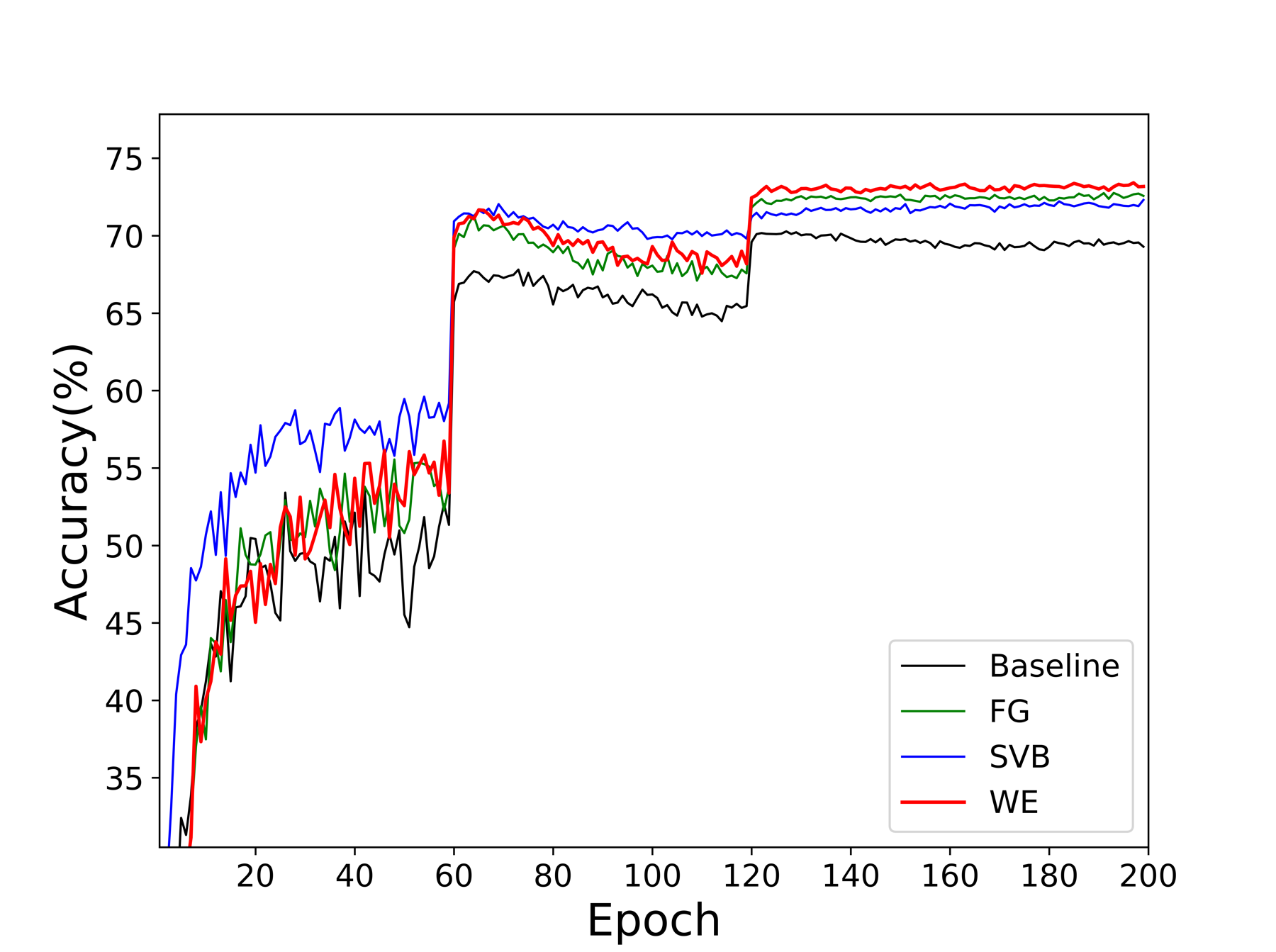}}
	\caption{Convergence curves of the compared methods}
	\label{fig:data_distribution}
	\vspace{0.2in}
\end{figure*}

The convergence cures are illustrated in Figure \ref{fig:data_distribution}. We can see that WE converges well and achieves the best performance in the last stages. WE begins to outperform FG in the second stage in all the cases. This shows that changing the filters more gently leads is more effective in reactivation.

Besides, we also report the additional running time compared to baseline for MobileNetV1 on CIFAR-10: WE(+1.2\%), FG-1(+2.4\%), RePr(+3.8\%), SVB(+5.3\%). This is consistent with the computational complexity analysis.

\begin{table}[h]
	\renewcommand{\arraystretch}{1.2}
	\caption{Classification accuracy (\%) comparison on CIFAR with lightweight networks}\label{CIFAR_lightweight}
	\scalebox{0.9}{
		\centering
		\begin{tabular}{llccc}
			\toprule
			Dataset         & Method     & MobileNetV1 & MobileNetV2 & ShuffleNetV1 \\
			\midrule
			
			\multirow{7}{*}{CIFAR-10}  & Baseline  &91.00 &92.63  &91.41  \\
			& FG-1   & 92.31            &94.18              &92.89 \\
			& FG-2 & 91.49 &93.38&92.40\\
			& FG-4 &92.39&93.89&92.86\\
			& SVB    &91.02          &94.31             &91.96\\
			& \textbf{WE}  &\textbf{92.44} & \textbf{94.36} &\textbf{93.04}\\
			\midrule
			\multirow{7}{*}{CIFAR-100} & Baseline    &68.41  & 72.27 &70.28\\
			& FG-1             & 70.30            & 75.10             &72.89\\
			& FG-2 &68.74&72.73&71.91\\
			& FG-4 &70.27&74.79&72.80\\
			& SVB           &64.86       &72.76              &72.39\\
			& \textbf{WE}  &\textbf{71.01} &\textbf{75.58}&\textbf{73.42}\\
			\bottomrule
	\end{tabular}}
\end{table}

\subsubsection{Result on ImageNet}
We demonstrate that WE also works well for the large-scale dataset by conducting experiment on ImageNet. Tables \ref{ImageNet_ResNet34} and \ref{ImageNet-MobileNet} show the results of ResNet-34 and MobileNetV2-0.5 \cite{DBLP:conf/cvpr/SandlerHZZC18}, respectively. Here, 0.5 denotes the width multiplier of  MobileNetV2. WE outperforms the filter-level reactivation methods FG-1 and RePr and the orthogonal weight method ONI. For lightweight networks, its improvement against baseline is more significant. We should emphasize that the extra computation brought by WE is negligible but it effectively improves performance.

\begin{table}[h]
	\renewcommand{\arraystretch}{1.1}
	\centering
	\caption{Comparion of weight evolution of ResNet-34 with other learning methods on ImageNet. }\label{ImageNet_ResNet34}
	\begin{tabular}{lcc}
		\toprule
		Architecture & Top-1 Acc. (\%) & Top-5 Acc. (\%) \\
		\hline
		Baseline         &72.60       & 90.91      \\
		FG-1     &73.29      &91.32       \\
		ONI      &72.96       &91.29       \\
		RePr      & 73.51      &--       \\
		\textbf{WE}  &  \textbf{73.56}       &  \textbf{91.38}      \\
		\bottomrule
	\end{tabular}
\end{table}
\begin{table}[h]
	\renewcommand{\arraystretch}{1.2}
	\centering
	\caption{Top-1 and Top-5 accuracy of MobileNetV2-0.5 on ImageNet. The image size is 128x128 for both training and testing.}\label{ImageNet-MobileNet}
	\begin{tabular}{lcc}
		\toprule
		Architecture        & Top-1 Acc. (\%) & Top-5 Acc. (\%) \\
		\hline
		Baseline            & 58.31       &81.38       \\
		\textbf{WE}   &  \textbf{60.91}      &  \textbf{83.39}      \\
		\bottomrule
	\end{tabular}
\end{table}

\subsection{Ablation study}
In order to better understand how weight evolution works, we conduct comprehensive ablation study experiment in this sub-section. All experiments in this sub-section are carried out with MobileNetV1 and ShuffleNetV1 on CIFAR-100.

\subsubsection{Global and local selection.}
We conducted experiments to investigate the effectiveness of our proposed global and local selection strategies.
The experiments are performed by turning global and local selection strategies on and off.
We denote the one with only global selection as WE-G, and the one with only local selection as WE-L.
The parameters $\hat{r}$ and $\gamma$ for WE-G and WE-L are fixed as 0.05 and 0.05, respectively.
We report the results in Table \ref{ablation_selection}.
The accuracy of both WE-G and WE-L are higher than the baseline, indicating that both strategies can select proper inferior filters for further weight reactivation. WE performs the best, which means that the two strategies are compatible to work together for better performance.

\begin{table}[h]
	\renewcommand{\arraystretch}{1.2}
	\centering
	\caption{Accuracy (\%) comparison of selection strategies.}\label{ablation_selection}
	\begin{tabular}{lcc}
		\toprule
		Method        & MobileNetV1 & ShuffleNetV1\\
		\hline
		Baseline         &68.41              &70.28\\
		WE-G &69.28             &72.81\\
		WE-L &69.53&72.32 \\
		\textbf{WE}  &\textbf{71.01}  &\textbf{73.42}\\
		\bottomrule
	\end{tabular}
\end{table}

\subsubsection{Forward matching strategy}
Compared to reverse matching in FG \cite{DBLP:conf/cvpr/MengCLXJSL20}, the proposed forward matching strategy keeps the importance order of both inferior and dominant filter sets. Here we show the superiority of forward matching. We use WE-RM to denote weight evolution with reverse matching. The results are given in Table \ref{ablation_forward}. We can see that the performance improvement of WE over WE-RM is significant. Note that compared to baseline, the reverse matching strategy improves the performance, which also verifies the effectiveness of selection and crossover strategies again.
\begin{table}[h]
	\renewcommand{\arraystretch}{1}
	\centering
	\caption{Accuracy (\%) comparison of froward and reverse matching strategies.}
	\label{ablation_forward}
	\begin{tabular}{lcc}
		\toprule
		Method        & MobileNetV1 & ShuffleNetV1\\
		\hline
		Baseline         &68.41              &70.28\\
		WE-RM     & 70.71             &72.38\\
		\textbf{WE}  &\textbf{71.01}  &\textbf{73.42}\\
		\bottomrule
	\end{tabular}
\end{table}

\subsubsection{Adaptive coefficient $\alpha$.}
\begin{figure}[h]
	\centering
	\subfigure[ShuffleNetV1  CIFAR-100]{\includegraphics[width=0.23\textwidth]{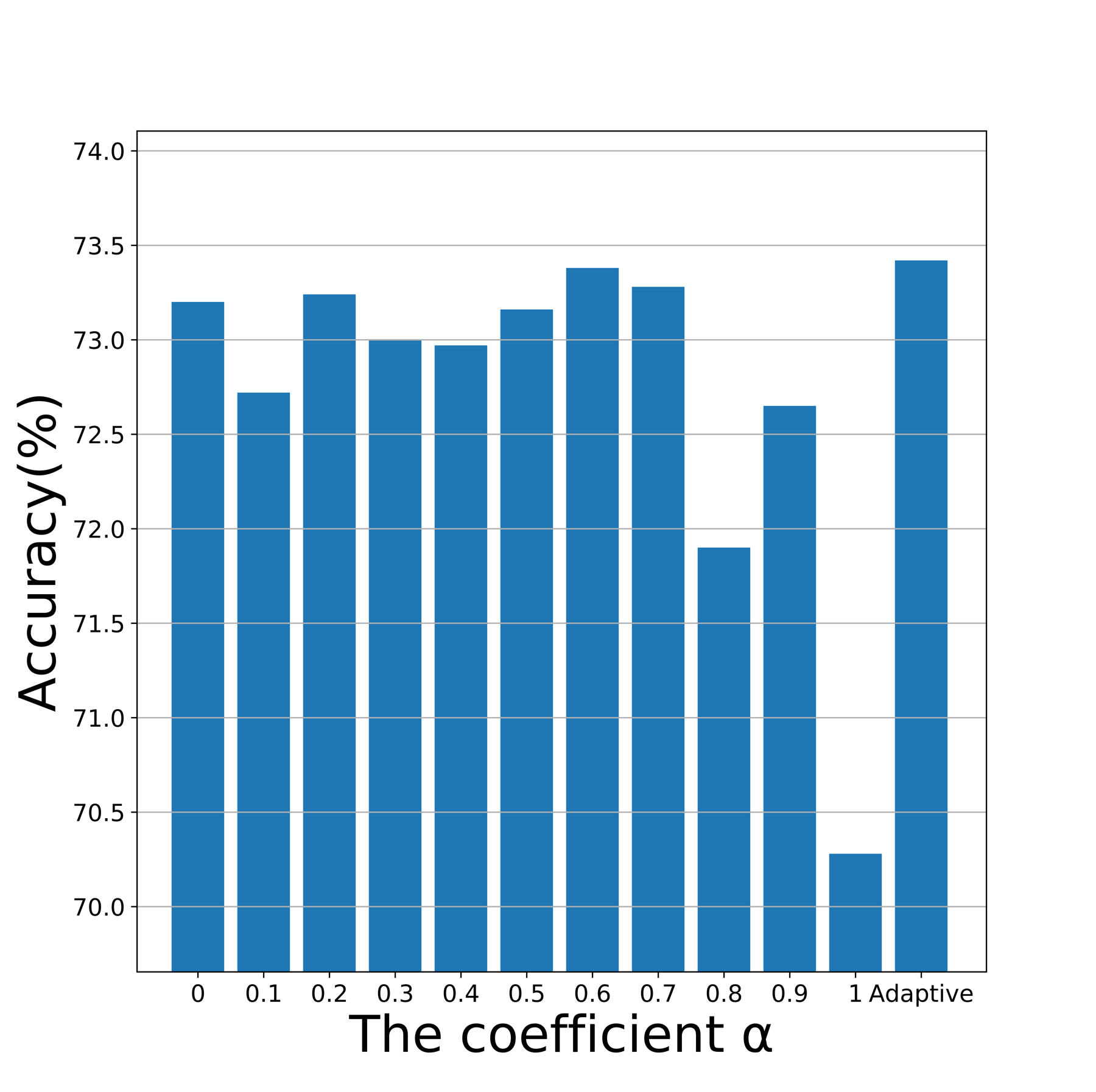}}
	\subfigure[MobileNetV1  CIFAR-100]{\includegraphics[width=0.23\textwidth]{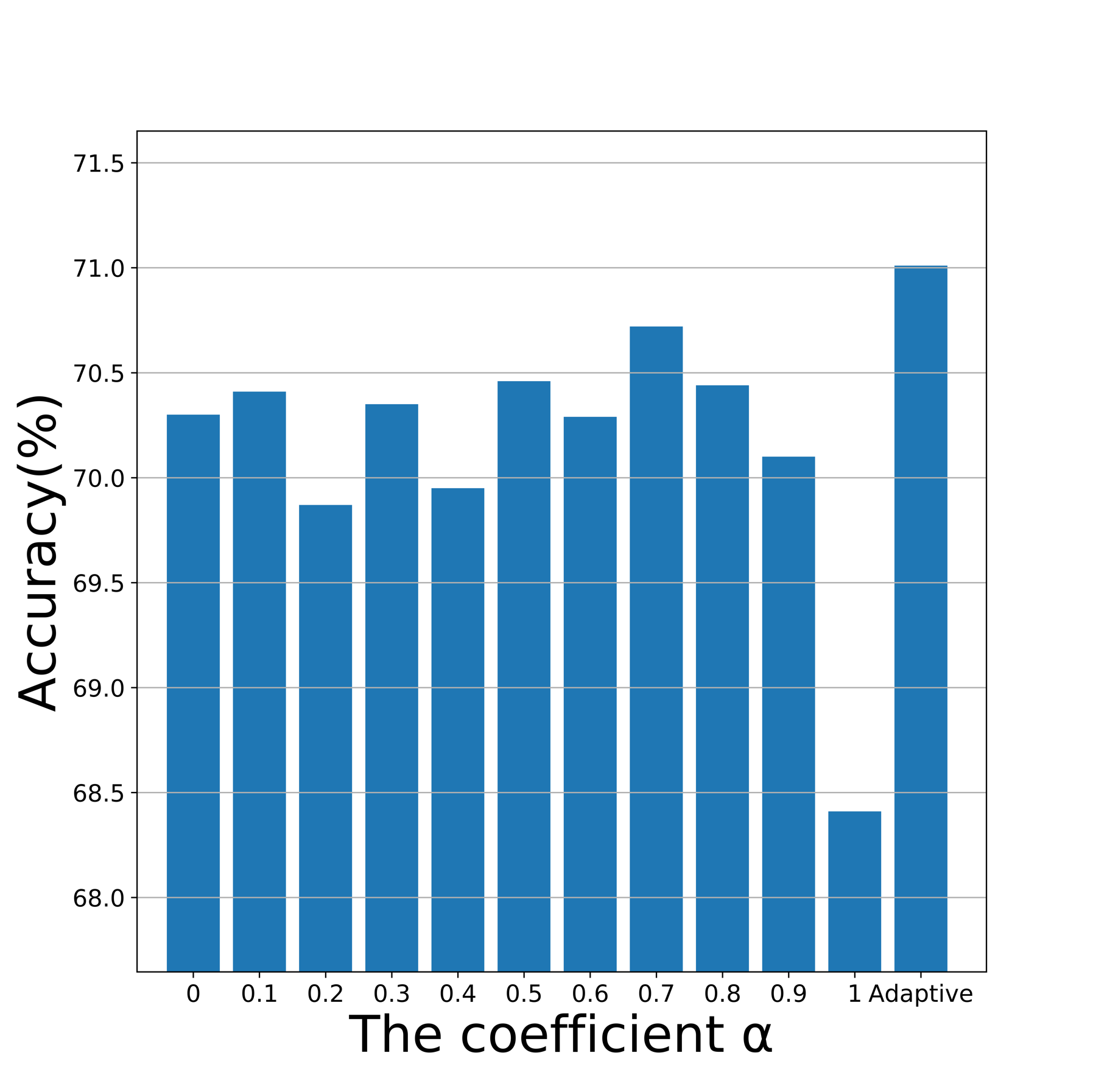}}
	\caption{Effectiveness of adaptive coefficient $\alpha$}\label{ablation_alpha}
	\vspace{0.02in}
\end{figure}

The crossover strategy mainly depends on the coefficient $\alpha$. In order to have a deeper understanding of the adaptive setting method, we compare the performance of using different fixed values of $\alpha$ and adaptive $\alpha$. For the fixed values, we vary them from 0 to 1 with a step size of 0.1, and the comparison results are shown in Figure \ref{ablation_alpha}. When $\alpha=1$, WE degenerates to the baseline. The performance is significantly improved even when $\alpha=0.9$, i.e., only a small portion of the dominant weight is utilized. This shows that using the dominant weight for reactivation is very effective. When the inferior weight is directly replaced by the dominant weight ($\alpha=0$), the performance is better than the baseline but not the best. This is because the inferior weight is obtained through network training and still carries some useful information. Thus, taking a good balance between the inferior and dominant weights is very important. For both networks, adaptive coefficients achieve the best performance, which indicates that our adaptive setting method is able to make a good trade-off.

\subsubsection{Element-Level Reactivation.}
To verify that finer level structure reactivation works better, we perform filter-level reactivation on the inferior filters selected by both selection strategies for comparison. We use the weighted sum of inferior and dominant filters to replace the original inferior filter with adaptive coefficient computed by relative importance of filters' $\ell_{1}$ norms similar to Eq. (\ref{alpha}).
We denote this comparison method as filter-level-GL. We report the results in Table \ref{ablation_element}. We can see that element-level reactivation outperforms filter-level reactivation.

\begin{table}[h]
	\renewcommand{\arraystretch}{1}
	\centering
	\caption{Accuracy (\%) comparison of element-level and filter-level reactivation}\label{ablation_element}
	\begin{tabular}{lcc}
		\toprule
		Method        & MobileNetV1 & ShuffleNetV1\\
		\hline
		Baseline         &68.41              &70.28\\
		Filter-level-GL     & 70.37             &73.10\\
		\textbf{WE}  &\textbf{71.01}  &\textbf{73.42}\\
		\bottomrule
	\end{tabular}
\end{table}

\subsubsection{Reactivation of BN layer and Convolutional Layer.}
Here we study the effect of reactivating convolution layers and BN layers separately.
As shown in Table \ref{ablation_BN}, WE without BN or WE without CONV can achieve higher accuracy than the baseline, but their accuracy is lower than WE. This demonstrates that WE works for both kinds of layers and their combination makes better.
\begin{table}[h]
	\renewcommand{\arraystretch}{1}
	\centering
	\caption{Accuracy (\%) comparison of WE that does not deal with BN layer or the convolution layer.}
	\label{ablation_BN}
	\begin{tabular}{lcc}
		\toprule
		Method        & MobileNetV1 & ShuffleNetV1\\
		\hline
		Baseline         &68.41              &70.28\\
		WE without BN     & 69.45             &70.57\\
		WE without CONV     & 69.72             &72.37\\
		\textbf{WE}  &\textbf{71.01}  &\textbf{73.42}\\
		\bottomrule
	\end{tabular}
\end{table}

\section{Conclusion}
In this paper, we propose WE that reactivate inferior weight during training. Different from previous reactivation methods that operate on the entire filter, WE focuses on the finer structure and shows that weight element reactivation performs better than filter reactivation.  WE is a plug-in training method and consistently improves the performance of existing network architecture on benchmark image classification datasets. Especially, the performance gain is more significant for lightweight networks.

\section{Acknowledgments}
This work is supported by the Natural Science Foundation of Guangdong Province, China (2018A030313474), the National Natural Science Foundation of China (61802131, U1801262, and 61972163), the Natural Science Foundation of Guangdong Province, China (2019A15\\15012146 and 2020A1515010781), Key-Area Research and Development Program of Guangdong Province, China (2019B010154003), the Guangzhou Key Laboratory of Body Data Science (201605030011), Science and Technology Project of Zhongshan (2019AG024), the Fundamental Research Funds for the Central Universities, SCUT (2019PY21 and 2019MS028), and China Southern Power Grid Company Limited Science and technology project funding (GZHKJXM202\\00058).

\bibliographystyle{ACM-Reference-Format}
\bibliography{sample-base}

\end{document}